\documentclass[letterpaper, 10 pt, conference]{ieeeconf}  
\usepackage{graphicx}
\usepackage{mdwlist}
\usepackage{fmtcount}
\usepackage{algorithmic}
\usepackage{algorithm} 
\usepackage{array}                                                         
\usepackage[tight,footnotesize]{subfigure}                                                          

\IEEEoverridecommandlockouts                              
\title{\LARGE \bf
Computer Aided Detection of Anemia-like Pallor
}

\author{Sohini Roychowdhury$^{1*}$, Donny Sun$^{1}$, Matthew Bihis$^{1}$, Johnny Ren$^{1}$, Paul Hage$^{1}$, Humairat H. Rahman$^{2}$
\thanks{$^{1}$Department of Electrical and Computer Engineering, University of Washington, Bothell, WA-98011. $^{*}$Email: roych@uw.edu.}
\thanks{$^{2}$Department of Environmental and Occupational Health, University of South Florida, Tampa, FL-33612}
}
\begin{document}

\maketitle
\thispagestyle{empty}
\pagestyle{empty}

\begin{abstract}
Paleness or pallor is a manifestation of blood loss or low hemoglobin concentrations in the human blood that can be caused by pathologies such as anemia. This work presents the first automated screening system that utilizes pallor site images, segments, and extracts color and intensity-based features for multi-class classification of patients with high pallor due to anemia-like pathologies, normal patients and patients with other abnormalities. This work analyzes the pallor sites of conjunctiva and tongue for anemia screening purposes. First, for the eye pallor site images, the sclera and conjunctiva regions are automatically segmented for regions of interest. Similarly, for the tongue pallor site images, the inner and outer tongue regions are segmented. Then, color-plane based feature extraction is performed followed by machine learning algorithms for feature reduction and image level classification for anemia. In this work, a suite of classification algorithms image-level classifications for normal (class 0), pallor (class 1) and other abnormalities (class 2). The proposed method achieves 86\% accuracy, 85\% precision and 67\% recall in eye pallor site images and 98.2\% accuracy and precision with 100\% recall in tongue pallor site images for classification of images with pallor. The proposed pallor screening system can be further fine-tuned to detect the severity of anemia-like pathologies using controlled set of local images that can then be used for future benchmarking purposes. 
\end{abstract}


\IEEEpeerreviewmaketitle

\section{Introduction}
Paleness or pallor is a manifestation of anemia, defined as abnormally low hemoglobin concentrations in the blood, which can be caused by blood loss, malnutrition, or other pathologies. Hemoglobin serves a vital role in the blood, carrying oxygen to the tissues. While anemia affects around 1.6 billion people worldwide, it is known to affect women and preschool children 5-8 times more than men \cite{ref2}. The Centers for Disease Control and Prevention estimate that the number of visits to emergency departments in North America with anemia as the primary hospital discharge diagnosis have been steadily increasing from 1990-2011 \cite{CDC}. Chronic anemia can contribute to problems such as chronic fatigue, or more severe problems such as heart failure, limb ischemia, and pregnancy complications. With such large percentages of the present-day population at risk of the detrimental impacts of anemia, the design of computer aided diagnostic (CAD) systems that can screen patients with anemia from the normal patients by detecting pallor non-invasively becomes necessary. CAD and point of care (POC) applications are aimed at providing quick ``expert'' diagnostics for screening and resourcefulness of treatment and care-delivery protocols. Also, such systems are  useful especially in a telemedicine paradigm, where, the patient and the care-provider may not be geographically co-located. Facial images have been extensively useful for security, authentication, identification and expression detection purposes \cite{Michelle}. This work is aimed at utilizing facial pallor site images for anemia-like medical screening applications.

This paper makes three key contributions. First, spatial, color-based and gradient based features are analyzed to detect the optimal combination for prediction of patient pallor. We observe that Frangi-filtering and gradient filtering enhance the image separability for pallor severity in eye and tongue pallor site images, respectively. Second, a hierarchical classification strategy is proposed using an optimal set of color-based and intensity-based features for screening normal, anemic and abnormal pallor site images. We observe 72-86\% separability of normal from abnormal images for eye and tongue pallor site images, respectively. Third, the discriminating contribution of each pallor site image for anemia-like pallor is analyzed. Here, we observe that the eye pallor site has higher area under Receiver Operating Characteristic curves (AUC) when compared to that of the tongue pallor site images.

\section{Materials and Methods}\label{2}
Based on prior works that analyze pallor site images for anemia-like diagnosis \cite{validity}, this work focusses on the eye images with visible conjunctiva and tongue images for anemia-like pallor detection. There are two primary objectives of the analysis presented in this work. First, the color and region-based features in each pallor site image are analyzed to determine the most discriminating features for pallor classification. Second, the importance of the eye and tongue pallor sites are assessed to identify the most significant pallor site for anemia-like pathology classification. A description of the image data sets under analysis and the proposed methods are given below.

\subsection{Data}
\begin{figure}[ht]
	\begin{center}
		\
		\includegraphics[width = 2.9in, keepaspectratio=true]{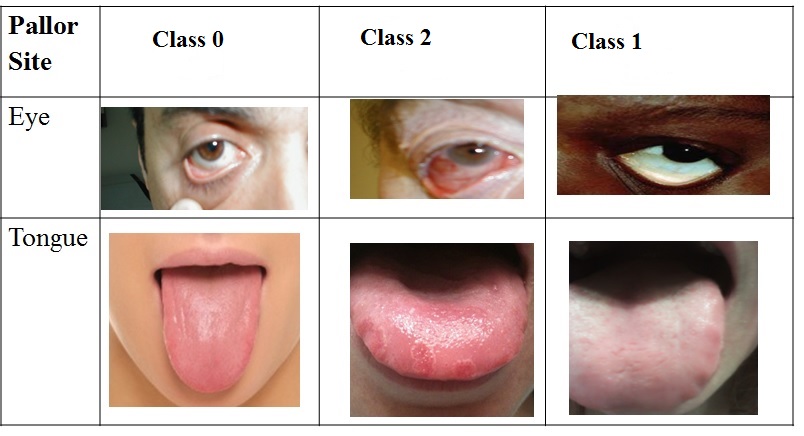}
		\caption{Example images of eye and tongue pallor sites with varying pallor severity grades.}     \label{sample}
	\end{center}
\end{figure}

A set of 27 eye images and 56 tongue images are collected and manually annotated for subjective pallor indices. These images represent uncontrolled imaging conditions and a wide variety in patient demographics. Each pallor site image has dimensions ranging from [155x240] to [960x1280] pixels with storage size of 8kB to 252kB per image. Also, every pallor site image that is gathered from public domain sources, is manually annotated for pallor severity grade. While grade 0 refers to normal patients, grade 1 refers to patients with anemia-like pathologies and grade 2 is indicative of pathologies/abnormalities associated with the specific pallor site that is not anemia-like. Examples of sample images corresponding to each severity grade from the eye and tongue pallor site images are shown in Fig. \ref{sample}. The goal of the overall automated system is to classify each pallor site image with output class label [0, 1, 2], representative of patient's anemia-like pallor.

For the eye and tongue pallor site images, the numbers of images belonging to the sample class labels representative of severity [0,1,2] are [6, 7, 14], and [18, 3, 35] respectively. For homogeneous processing purposes, each image ($I$) is resized to [125x125] pixels each. For automated pattern recognition from the pallor site images, the eye and tongue data sets are partitioned into training and test data sets, respectively. Due to the limited data sizes, feature learning and data modeling is performed using 5-fold cross validation for the eye images and 3-fold cross validation for tongue images \cite{cherkassky}. This is to ensure similar proportions of class sample data in each of the folded data sets.

\subsection{Proposed Methods}\label{3}
Since analysis of pallor site images for anemia-like pathologies is a novel idea, there are no existing methods in literature. However, based on the variabilities introduced by the data sets, two data models are analyzed for pallor severity grade classification tasks. The first model ($M1$) is designed to detect specific spatial regions of interest (ROIs) that are indicative of patient pallor. In this model, the pallor site images are segmented to extract several ROIs, followed by extraction of pixel intensity features corresponding to color plane and gradient images within the ROIs. Next, the intensity-based features are ranked to detect the most discriminating set of features from the training data set that ensure maximum accuracy in the validation data set \cite{AGMIC}\cite{cherkassky}. Finally, the most optimal feature set is utilized for pallor severity grade classification.

The second model ($M2$) is designed to detect the most significant color planes and gradient images for pallor classification. For each pre-processed color plane image, a mask `$g$' of the eye or tongue region is detected.  Color plane transformations are then applied to each pallor-site image, thus resulting in the following 12 image planes: red, green, blue, hue, saturation, intensity (from RGB to HSV transformation), lightness, a-color plane, b-color plane (from RGB to Lab transformation), luminance, 2 chrominance planes (from RGB to Ycbcr transformation). Next the first order gradient filtered image in horizontal and vertical directions is extracted from each color image plane ($I^{G}$) and superimposed on the color image plane itself, thereby resulting in 12 additional images. Finally, each color image plane is Frangi-filtered \cite{neo} to extract the second order edges ($I^{F}$) and superimposed on the image itself, generating 12 additional images per pallor site image. Using this process, 36 color and edge enhanced images are extrapolated per pallor site image. Using cross-validation method, the 36 extrapolated images are ranked to identify the most discriminating image color and edge enhancement procedure for pallor classification.

\subsubsection{Image Segmentation}
The first step for model $M1$ involves spatial segmentation of the pallor site image into several ROIs. For the eye pallor site images, the sclera and conjunctiva regions while for the tongue pallor site images the inner and outer tongue regions would constitute the different ROIs. For segmentation of eye images, the first step is detection of the iris, followed by detection of the surrounding sclera, followed by conjunctiva detection. The steps for detecting $R_{iris}, R_{sclera}, R_{conj}$ as the iris, sclera and conjunctiva regions, respectively are shown in (\ref{eq_eye})-(5). First, the scaled red plane image in [0,1] pixel range is thresholded to detect dark regions with area greater than 100 pixels only in (1). These regions are represented by $R$. The $R_{iris}$ is the dark region with the most elliptical shape (i.e., highest ratio of major axis length ($\phi$) and minor axis lengths ($\psi$) in (2). Next, the green plane image within the masked $g$ region is subjected to watershed transformation using circular structuring element ($se$) of radius 5 in (3). This results in several sub-region segmented image $W$. The iris region is removed from the image $W$ followed by detection of the remaining sub-regions in $W$ that intersect with the edge of $R_{iris}$ in (4). Since the sclera region lies right outside the iris, the edges of the sclera region and the iris region intersect. Finally, the conjunctiva region is detected as the remaining regions in mask $g$ after removing the iris and sclera regions in (5). For the tongue pallor site images, the masked green plane image within masked region $g$ is subjected to watershed transformation, thereby resulting in image $W$ with several sub-regions $R$. Next, the outer edge of the tongue is detected in image $E$ using the `Sobel' filter. The sub-regions in $R$ that intersect with the outer tongue edge regions represent the outer regions in the tongue ($R_{outer}$). The remaining regions in the $R$ after removing the $R_{outer}$ regions represent the inner tongue regions ($R_{inner}$).

\begin{eqnarray}\label{eq_eye}
R\leftarrow (I_{red}<0.1), Area(R)>100.\\
R_{iris}\leftarrow \arg_{R}\max \frac{\phi_R}{\psi_R}.\\
W\leftarrow Watershed[I_{green}\circ g,se].\\
R_{sclera}\leftarrow (W-R_{iris})\cap edge(R_{iris}).\\
R_{conj}=g-[R_{iris}+R_{sclera}].\\ \nonumber
\end{eqnarray}

\subsubsection{Color Planes and Gradient Feature Extraction}
\begin{figure}[ht]
	\begin{center}
	 \subfigure[]{\includegraphics[width = 3.0in, height=1.8in]{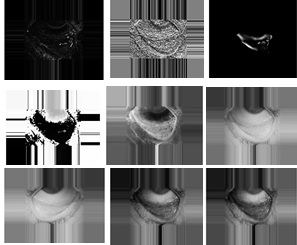}}	
		\subfigure[]{\includegraphics[width = 3.0in, height=1.8in]{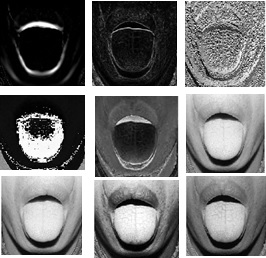}}	
		\caption{Examples of color and gradient plane images. Top row: $I^{G}_{green}$ magnitude, direction, $I^{F}_{green}$. Middle row: red, green, blue color planes. Bottom row: hue, saturation, intensity color planes for (a) Eye. (b) Tongue images, respectively.}     \label{feat}
	\end{center}
\end{figure}

For model $M1$, 54 features are extracted per image using pixel intensity-features from color and gradient transformed images from various segmented sub-regions in each image as shown in Fig. \ref{feat}. For the eye images, 27 features are extracted for the sclera and conjunctiva region each, corresponding to the max, mean, variance of pixels in the following image planes: red, blue, green, hue, saturation, intensity, $I^{G}_{green}$, $I^{F}_{green}$. For the tongue images, 27 similar features are extracted for the inner and outer tongue regions, each.
\subsubsection{Classification}
The final step in data models $M1$ and $M2$  involve classification using a family of data models implemented on the Microsoft Azure Machine Learning Studio (MAMLS) platform for scalability. These classifiers, called Azure-based Generalized flow for Medical Image Classification (AGMIC) \cite{AGMIC}, involve grid-search based hyper-parameterization of several data models and selection of the optimal data model with highest classification accuracy on the validation data set. Here, automated parametrization of 8 sets of data models is performed \cite{AGMIC} including support vector machines, logistic regression, boosted decision tree, decision forest, decision jungle, neural networks, Poisson regression and k-nearest neighbors. At the the end of the training process, one optimal data model with lowest classification error is selected and used for classification of the test data samples. 
\section{Experiments and Results}\label{4}
The images from tongue and eye data sets are analyzed separately. Since the data sets contain 3 classes of data samples, 2-step hierarchical classification is performed \cite{dream} for data separability analysis. In the first hierarchical step, the normal images are separated form the abnormal ones (class 0 vs. class 1, 2) or images with anemia are separated from the non-anemic ones (class 1 vs. class 0, 2). In the second hierarchical step, the remaining class samples are separated, i.e., (class 1 vs. class 2) or (class 0 vs. class 2), respectively.

Three categories of experiments are performed to identify the most discriminating set of intensity-based, spatial, and color-based features useful for classification of pallor severity grade. First, the intensity-based features extracted per image using model $M1$ are subjected to feature ranking followed by double cross-validation \cite{AGMIC} to identify the optimal set of intensity-based features. Second, the color-plane transformations applied in model $M2$ are analyzed to identify the most significant spatial and color-planes. Third, the optimal intensity-based features are used for classification in model $M1$ and the optimal color planes are used to classify the images using model $M2$. 
\subsection{Intensity-based Feature Learning}
The 54 intensity-based features extracted per pallor site image in model $M1$ are ranked using the F-score, Mutual Information and Chi-squared scoring methods \cite{AGMIC} and multi-class classification. We observe that for both the eye and tongue data sets, the 27 intensity-based features corresponding to the color planes, gradient and Frangi-filtered images from the conjunctiva region and the inner tongue regions, respectively, are optimal for classification of normal patients from abnormal ones. However, all the 54 intensity-based features corresponding to the conjunctiva and sclera regions in the eye and the inner and outer regions in the tongue are significant for classification of anemic images from abnormal ones. This observation is inline with the domain knowledge regarding the appearance of the conjunctiva in eye and inner tongue regions for identifying normal patients and analysis of all regions in the eye and tongue for further detection of abnormalities.

\subsection{Color-plane based Feature Learning}
The 36 color and gradient planes extrapolated per pallor site image using model $M2$ are analyzed for multi-class classification performances. For the eye data set with 27 images, [36x27=972 images] and for the tongue data set with 56 images, [36x56=2016 images] are subjected to classification. The rate of correct classification for each color and gradient plane image is analyzed to identify the most discriminating planes. We observe that for the eye data set, images ($I^{F}_{hue}$) and ($I^{F}_{sat}$) result in the maximum classification accuracy of 56\%. For tongue images, lightness and a-channel planes ($I^{G}_{L}, I^{G}_{a}$) achieve maximum classification accuracy of 65\%.

\subsection{Classification Performance Analysis}
For the eye data set, model $M1$ with AGMIC flow is implemented with 27 intensity-based features from conjunctiva region for step-1 of hierarchical classification followed by 54 intensity-based features from sclera and conjunctiva regions for step-2 of hierarchical classification, respectively. The classification performance of models $M1$ and $M2$ on the eye images are shown in Table \ref{tab1}. Here, we observe that the $M1$ model implemented with decision forest data model has the best image classification performance. 

\begin{table}[ht]
\caption{Performance Analysis of Pallor classification Models on Eye Images.}
\begin{center}
\begin{tabular}{|c| c c |c c|}
\hline
Model&M1,AGMIC&&M2,AGMIC&\\ \hline
Task&0/1,2&1/2&0/1,2&1/2\\ \hline  \hline
PR&0.85&0.57&0.45&0.42\\ \hline
RE&0.67&0.8&0.57&0.42\\ \hline
Acc&0.86&0.67&0.53&0.74\\ \hline
AUC&0.75&0.675&0.41&0.49\\ \hline
 \end{tabular}
 \label{tab1}
\end{center}
\end{table}

For the tongue data set, model $M1$ with AGMIC flow is implemented with 27 intensity-based features from inner tongue region for step-1 of hierarchical classification followed by 54 intensity-based features from inner and outer tongue regions for step-2 of hierarchical classification, respectively. The classification performance of models $M1$ and $M2$ on the tongue images are shown in Table \ref{tab2}. Here, we observe that the $M2$ model implemented with boosted decision trees data model has the best screening performances. 
\begin{table}[ht]
\caption{Performance Analysis of Pallor classification Models on Tongue Images.}
\begin{center}
\begin{tabular}{|c| c c |c c|}
\hline
Model&M1,AGMIC&&M2,AGMIC&\\ \hline 
Task&1/0,2&0/2&0/1,2&1/2\\ \hline \hline
PR&0.982&0.51&0.8&0.77\\ \hline
RE&1&0.53&0.81&0.87\\ \hline
Acc&0.982&0.61&0.72&0.73\\ \hline
AUC&0.83&0.574&0.78&0.67\\ \hline
 \end{tabular}
 \label{tab2}
\end{center}
\end{table}

\section{Conclusions and Discussion} \label{5}
In this work, we present a variety of image-based feature extraction, segmentation and data modeling approaches for classification and screening of anemia-like pallor using focused facial pallor site images. We perform three categories of experiments on eye and tongue pallor site images that are acquired from the public domain. The first category of experiments demonstrates that image intensity-based features corresponding to some specific spatial ROIs are significant for separating normal images from abnormal ones that must be further analyzed by specialists. Here, we find that the conjunctiva region in the eye and the inner tongue regions are significant for identification of normal images and abnormal images from eye and tongue pallor site images, respectively. The second category of experiments detects the most discriminating color and gradient plane-transformed images that are significant for classification of image-based pallor. This experiment demonstrates that Frangi-filtered hue and saturation color planes and first-order gradient filtered luminance channel planes are most significant for pallor classification using eye and tongue images, respectively. Our analysis leads to detection of intensity-based features from conjunctiva region in the hue and saturation color planes superimposed with Frangi-filtered edges for optimal separation of normal images from anemic or abnormal images using the eye pallor site images. Also, intensity-based features from the inner tongue regions in the luminance color planes superimposed with gradient filtered edges are significant for classification of abnormal images from normal and anemic ones using the tongue pallor site images. In the third category of experiments, we observe that the image segmentation and classification results in 86\% screening accuracy for eye images while color-transformations and gradient filtering leads to 98\% screening accuracy for tongue images. Thus, the proposed system is capable of severity screening for anemia using facial pallor site images in under 20 seconds of computation time per image. 

Future works will be directed towards analysis of additional data sets acquired under controlled imaging conditions. Since the data sets under analysis in this work represent a huge variety of imaging condition variabilities, the observations from the experimental analysis are more generalizable yet limited in classification capabilities. Future efforts will be directed towards correlation of the automated pallor severity grade obtained from the facial pallor site images with respect to the actual patient hemoglobin count for pre-clinical evaluations.

\bibliographystyle{IEEEtran}
\bibliography{IEEEabrv,references}

\end{document}